\documentclass[runningheads]{llncs}
\usepackage{graphicx}
\usepackage{amsmath,amssymb} 
\usepackage{color}
\usepackage[dvipsnames]{xcolor}
\usepackage[width=122mm,left=12mm,paperwidth=146mm,height=193mm,top=12mm,paperheight=217mm]{geometry}

\usepackage{xspace}
\usepackage{hyperref} 
\usepackage{cleveref}
\usepackage{multirow}
\usepackage{array}

\newcommand{\sota}{state-of-the-art\xspace}

\newcommand{\bw}{\mathbf{w}}
\newcommand{\bx}{\mathbf{x}}

\newcommand{\iou}{\operatorname{IoU}}
\newcommand{\siou}{\operatorname{SIoU}}
\newcommand{\boxsim}{\rho}
\newcommand{\lnbreakcell}[2][c]{%
  \begin{tabular}[#1]{@{}c@{}}#2\end{tabular}}

\newcommand{\myparagraph}[1]{\vspace{0.25em}\noindent {\bf #1.}}
\newcommand{\ie}{{\it i.e.\xspace}}
\newcommand{\eg}{{\it e.g.\xspace}}
\newcommand{\etal}{\mbox{\emph{et al.\xspace}}}

\begin{document}
\pagestyle{headings}
\mainmatter
\title{Learning the semantic structure of objects \\ from Web supervision} 

\author{}
\institute{}

\maketitle

{\centering
David Novotny$^1$ ~ ~ Diane Larlus$^2$ ~ ~ Andrea Vedaldi$^1$\\
}

\vspace{0.2cm}

\begin{minipage}{.45\textwidth}
\centering
$^1$\small{Visual Geometry Group\\University of Oxford\\}
{\ttfamily \{david,andrea\}@robots.ox.ac.uk}
\end{minipage} 
\begin{minipage}{.45\textwidth}
\centering
$^2$\small{Computer Vision Group\\Xerox Research Centre Europe\\}
{\ttfamily diane.larlus@xrce.xerox.com}
\end{minipage}

\begin{abstract}

While recent research in image understanding has often focused on recognizing
\emph{more types of objects}, understanding \emph{more about the objects} is
just as important. Recognizing object parts and attributes has been extensively
studied before, yet learning large space of such concepts remains elusive due to
the high cost of providing detailed object annotations for supervision. The key
contribution of this paper is an algorithm to learn the \textit{nameable parts}
of objects automatically, from images obtained by querying Web search
engines. The key challenge is the high level of noise in the annotations; to
address it, we propose a new unified embedding space where the appearance and
geometry of objects and their semantic parts are represented uniformly. Geometric
relationships are induced in a soft manner by a rich set of non-semantic
mid-level anchors, bridging the gap between semantic and non-semantic parts. We
also show that the resulting embedding provides a visually-intuitive
mechanism to navigate the learned concepts and their corresponding
images.
\end{abstract}

\section{Introduction}\label{s:intro}

Modern deep learning methods have dramatically improved the performance of computer vision algorithms in selected tasks such as image
classification~\cite{krizhevsky12imagenet} and object detection~\cite{girshick2014rich}. Parallel advances in tasks such as image
captioning~\cite{frome13devise,karpathy14deep}, activity recognition~\cite{simonyan14two-stream}, and many others have ventured far beyond classification
and detection in order to extract richer information from visual scenes. Even so, image understanding remains rather crude, oblivious to most of the nuances of
real world images.
Consider for example the notion of \textit{object category}, which is a basic unit of understanding in computer vision. Modern benchmarks consider an
increasingly large number of such categories, from thousands in the ILSVRC challenge \cite{russakovsky15imagenet} to hundred thousands in the full
ImageNet \cite{deng09imagenet}. However, there is only limited understanding of their internal semantic structure and geometry.
 
In this paper we aim at filling this gap by jointly \emph{learning about objects, their semantic parts, and their geometric relationship}. Semantic
\textit{nameable} parts play a crucial role in visual understanding. However, learning them on a large scale using standard methods faces the difficulty of collecting vast quantities of corresponding annotated example images. Instead, scalable algorithms must be designed to discover this information, with \emph{minimal or no supervision}.

\begin{figure}[t!]
  \centering
  \includegraphics[width=1\columnwidth]{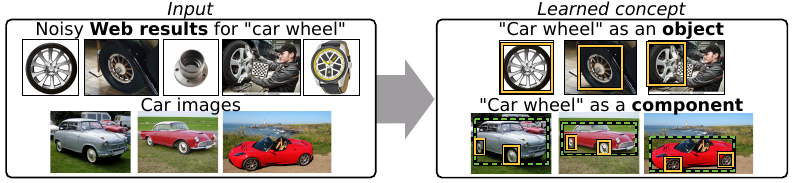}  
  \caption{Our goal is to learn the semantic structure of objects automatically
    using Web supervision. For example, given noisy images obtained by querying
    an Internet search engine for ``car wheel'' and for ``cars'',
    we aim at learning the ``car wheel'' concept, and its \textbf{dual
      nature}: as an object in its own right, and as a component of another
    object.}\label{f:splash}
\end{figure}

\begin{figure}[t!]
\begin{center}
  \includegraphics[width=\linewidth]{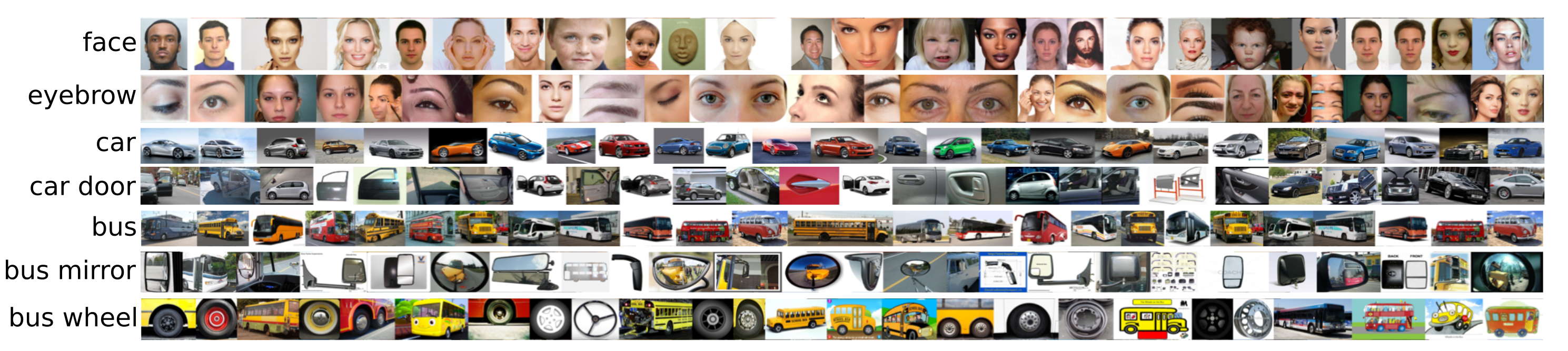}
\end{center}
\vspace{-2.5em}
\caption{Top images retrieved from an Internet search engine for some example queries. Note that part results are more noisy
  than full object results (the remaining collected images get even noisier, not shown here). \label{fig:query}}
\end{figure}

As others have done for the problem of learning visual objects, in this paper we look at Web supervision to learn object parts from thousands of images obtained automatically by querying search engines (\textit{crf}.~\cref{f:splash}). However, this poses two significant challenges: identifying images of the parts in very noisy Web results (crf.~\cref{fig:query}) while, at the same time, bridging the scale difference between parts seen in the context of the whole object or in isolation. The latter suggests in fact that \textbf{parts have a dual nature}: as components of an object as well as objects in their own right (\cref{f:splash} right), and models should be able to capture both. In order to address such challenges, we propose a new method to reason robustly about visual concepts and their geometric relationships.

Our first idea is to use the \emph{same representation for both objects and parts}, thinking them as generic ``semantic visual entities''.  Differently from methods such as Deformable Part Models (DPM)~\cite{fischler73the-representation,felzenszwalb10object}, our representation does not differentiate between objects and subordinate parts, promoting flexibility and robustness. Our second idea is to \emph{leverage non-semantic parts to learn about semantic ones}; methods such as DPMs seek in fact visually stable parts, that are often non-semantic. While these are not very interesting for semantic abstraction, they may provide reliable geometric anchors to represent object deformations.

These two ideas come together in the two main contributions of the
paper. The first contribution (\cref{s:mil}) is a novel embedding that captures
appearance and geometry of all visual entities, either objects or semantic
parts, in the same space. Geometry is expressed robustly against an
object-centric reference frame implicitly captured by non-semantic anchor
parts. The second contribution (\cref{s:mid-level}) is an effective method to
learn these non-semantic anchors, which is an alternative to significantly more complex part discovery methods.

A byproduct of our method is a large collection of images annotated with
objects, semantic parts, and their geometric relationships, that we refer to as
a \emph{visual semantic atlas} (\cref{s:exp-qualitative}). This atlas allows to \emph{visually navigate} images based on conceptual and geometric relations. It also emphasizes the dual nature of parts, as components of an object and as semantic categories, by naturally bridging images that zooms on a part or that contain the object as a whole.

\subsection{Related work}\label{s:related}

Our work touches on several active research areas: localizing objects with
weak supervision, learning with Web images, and discovering or learning
mid-level features and object parts.

\myparagraph{Localizing objects with weak supervision} When training models to localize objects or parts, it is impractical to expect large
quantities of bounding box annotations. Recent works have tackled the localization problem with only image-level annotations.  Among them, \textit{weakly
  supervised object localization} methods \cite{nguyen09weakly,pandey2011scene,deselaers12weakly,wang14weakly,hoffman14lsda,hoffman15detector,cinbis15weakly}
assume for each image a list of every object type it contains. In the \textit{co-detection}
\cite{joulin14efficient,tang14co-localization,ali14confidence,shi15bayesian} and \textit{co-segmentation}
\cite{joulin10efficient,vicente11object,joulin12multi,rubinstein13unsupervised} problems, the algorithm is given a set of images that all contain at least one
instance of a particular object.  They differ in their output: co-detection predicts bounding boxes, while segmentation predicts pixel-level masks. Yet,
co-detection, co-segmentation and weakly-supervised object localization (WSOL) are different flavors of the localization problem with weak supervision.
For co-detection and WSOL, the task is nearly always formulated as a multiple instance learning (MIL) problem
\cite{nguyen09weakly,pandey2011scene,song14slsvm,ali14confidence,li13harvesting}.  The formulation in \cite{hoffman14lsda,hoffman15detector} departs from MIL by
leveraging the strong annotations for some categories to transfer knowledge to the remaining categories. A few approaches model images using topic models
\cite{wang14weakly,shi15bayesian}.
Recently, CNN architectures were also proved to work well in weakly supervised scenarios \cite{bilen2015wsdnn}. 
We will compare with \cite{bilen2015wsdnn} in the experiments section.
None of these works have considered semantic parts.
Closer to our work, the method of \cite{cho15unsupervised} proposes unsupervised discovery of dominant objects using part-based region matching. Because of its
unsupervised process, this method is not suited to name the discovered objects or matched regions, and hence lack semantics. Yet we also compare with this approach
in our experiments.

\myparagraph{Learning from Web supervision}
Most previous works \cite{fergus05learning,parkhi12on-the-fly,schroff07harvesting,tsai11large} that learn from noisy Web images have focused on
image classification. Usually, they adopt an iterative approach that jointly learns models and finds clean examples of a target concept.
Only few works have looked at the problem of localization. Some approaches \cite{gunhee12on,rubinstein13unsupervised} discover common segments within a large set of Web
images, but they do not quantitatively evaluate localization. The recent method of \cite{chen15webly} localizes objects with bounding boxes, and evaluate the learnt models, but
as the previous two, it does not consider object parts.
Closer to our work, \cite{chen13neil} aims at discovering common sense knowledge relations between object categories from Web images, some of which correspond
to the ``part-of'' relation. In the process of organizing the different appearance variations of Webly mined concepts, \cite{divvala14learning} uses a
``vocabulary of variance'' that may include part names, but those are not associated to any geometry.

\myparagraph{Unsupervised parts, mid-level features, and semantic parts} Objects are modeled using the notion of \textit{parts} since the early work on
pictorial structure \cite{felzenszwalb03pictorialstructures}, in the constellation \cite{fergus03object} and ISM \cite{leibe09robust} models, and more recently
the DPM \cite{felzenszwalb10object}. Parts are most commonly defined as localized components with consistent appearance and geometry in an object. All these
works have in common to discover object parts without naming them. In practice, only some of these parts have an actual semantic interpretation.
\textit{Mid-level features} \cite{singh12unsupervised,doersch2013mid,Juneja13,endres13learning,doersch2015unsupervised,li2015mid} are discriminative \cite{endres13learning,bossard2014food} or rare
\cite{singh12unsupervised} blocks, which are leveraged for object
recognition. Again, these parts lack semantic. The non-semantic anchors that we
use share similarities with \cite{jian2015learning} and
\cite{doersch2015unsupervised}, that we discuss in \cref{s:mid-level}.
\textit{Semantic} parts have triggered recent interest \cite{zhang14finegrained,chen14detect,wang15joint}. These works require strong annotations in
the form of bounding boxes \cite{zhang14finegrained} or segmentation masks \cite{chen14detect,wang15joint} at the part level. Here we depart from existing work
and aim at mining semantic nameable parts with as little supervision as possible.

\section{Method}\label{s:method}

This section introduces our method to learn semantic parts using weak supervision from Web sources. The key challenge is
that search engines, when queried for object parts, return many outliers containing other parts as well, the whole object, or entirely unrelated things (\cref{fig:query}). In this setting, standard weakly-supervised detection approaches fail (\cref{s:experiments}). Our solution is a novel, robust, and flexible representation of object parts (\cref{s:mil}) that uses the output of a simple but very effective non-semantic part discovery algorithm (\cref{s:mid-level}).

\subsection{Learning semantic parts using non-semantic anchors}\label{s:mil}

\begin{figure}[t]
  \centering
\includegraphics[width=0.95\textwidth]{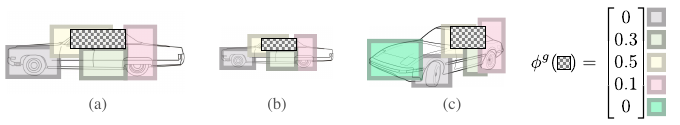}
\vspace{-1.8em}
\caption{{\bf Anchor-induced geometry.} (a) A set of anchors (light boxes) are obtained from a large number of unsupervised non-semantic part detectors. The geometry of a semantic part or object is then expressed as a vector $\phi^g$ of anchor overlaps. (b) The representation is scale and translation invariant. (c) The representation implicitly codes for multiple aspects.}\label{f:anchors}
\end{figure}

In this section, we first flesh out our method for weakly-supervised part learning and then dive into the theoretical justification of our choices.

\myparagraph{MIL: baseline, context, and geometry-aware} As standard in weakly-supervised object detection, our method starts from the \emph{Multiple Instance Learning} (MIL)~\cite{dietterich97solving} algorithm. Let $\bx_i$ be an image and let $\mathcal{R}(\bx_i)$ be a shortlist of image regions $R$ that are likely to contain objects or parts, obtained for instance using
selective search~\cite{uijlings13selective}. Each image $\bx_i$ can be either positive $y_i=+1$ if it is deemed to contain a certain part or negative $y_i=-1$ if not. MIL fits to this data a (linear) scoring function $\langle \phi(\bx_i|R),\bw\rangle$, where $\bw$ is a vector of parameters and $\phi(\bx_i|R)\in\mathbb{R}^d$ is a descriptor of the region $R$ of image $\bx_i$, by minimizing:
\begin{equation}\label{e:objective1}
\min_{\bw\in\mathbb{R}^d}
\frac{\lambda}{2}
\|\bw\|^2
+
\frac{1}{n} \sum_{i=1}^n \max\{0,
1 - y_i \max_{R\in\mathcal{R}(\bx_i)}\langle \phi(\bx_i|R),\bw\rangle\}
\end{equation}
In practice, \cref{e:objective1} is optimized by alternatively selecting the maximum scoring region for each image (also known as ``re-localization'') and optimizing $\bw$ for a fixed selection of the regions. In this manner, MIL should automatically discover regions that are most predictive of a given label, and which therefore should correspond to the sought visual entity (object or semantic part). However, this process may fail if descriptors are not sufficiently strong.

For \textbf{baseline MIL} the descriptor $\phi(\bx|R) =
\phi^a(\bx|R)\in\mathbb{R}^{d_a}$ captures the region's appearance. A common
improvement is to extend this descriptor with \textit{context information} by appending a descriptor of a region $R'=\mu(R)$ surrounding $R$, where $\mu(R)$ isotropically enlarges $R$; thus in \textbf{context-aware MIL}, $\phi(\bx|R) = \operatorname{stack}(\phi^a(\bx|R), \phi^a(\bx|\mu(R)))$.

Neither baseline or context-aware MIL leverage the fact that objects have a well-defined geometric structure, which significantly constrains the search space for parts. DPM uses such constraints, but as a fixed set of geometric relationships between part pairs that are difficult to learn when examples are extremely noisy. Furthermore, DPM-like approaches learn the most visually-stable parts, which often are \emph{not} the semantic ones.

We propose here an alternative method that captures geometry indirectly, on top of a rich set of unsupervised mid-level
non-semantic parts $\{p_1, ..., p_K\}$, which we call \emph{anchors} (\cref{f:anchors}). Let us
assume that, given an image $\bx$, we can locate the (selective search) regions $R_{p_k,\bx}$ containing each anchor $p_k$. We define the following geometric embedding $\phi^g$ of a region $R$ with respect to the anchors:
\begin{equation}\label{e:geomembed}
\phi^g(\bx|R) =
\begin{bmatrix}
\boxsim(R,R_{p_1,\bx}) \\
\vdots \\
\boxsim(R,R_{p_K,\bx})
\end{bmatrix}.
\end{equation}
Here $\boxsim$ is a measure such as intersection-over-union (IoU) that tells whether two regions overlap. By choosing a function $\boxsim$ such as IoU which is invariant to scaling, rotation, and translation of the regions, so is the embedding $\phi^g$. Hence, as long as anchors stay attached to the object, $\phi^g(\bx|R)$ encodes the location of $R$ relative to an object-centric frame. This representation is robust because, even if some anchors are missing or misplaced, the vector $\phi^g(\bx|R)$ is not greatly affected. The geometric encoding $\phi^g(\bx|R)$ is combined with the appearance descriptor  $\phi^a(\bx|R)$ in a joint appearance-geometric embedding
\begin{equation}\label{e:agkron}
    \phi^{ag}(\bx|R) = \phi^a(\bx|R) \otimes \phi^g(\bx|R)
\end{equation}
where $\otimes$ is the Kronecker product. After vectorization, this vector is used as a descriptor $\phi(\bx|R) = \phi^{ag}(\bx|R)$ of region $R$ in \textbf{geometry-aware MIL}. The next few paragraphs discuss its properties.

\myparagraph{Modeling multiple parts} Plugging
$\phi^{ag}$ of eq.~\eqref{e:agkron} into eq.~\eqref{e:objective1} of MIL results in the scoring function
 $\langle\bw, \phi^{ag}(\bx|R)\rangle = \sum_{k=1}^K \langle \bw_k, \phi^a(\bx|R)\rangle \rho(R,R_{p_k,\bx})$ which interpolates between $K$ appearance models \emph{based on how the region $R$ is geometrically related to the anchors $R_{p_k,\bx}$.} In particular, by selecting different anchors this model may capture simultaneously the appearance of all parts of an object. In order to control the capacity of the model, the smoothness of the interpolator can be increased by replacing IoU with a softer version, which we do next.

\myparagraph{Smoother overlap measure} The IoU measure is a special case of the following family of PD kernels  (proof in the appendix):
\begin{theorem}\label{t:overlap}
Let $R$ and $Q$ be vectors in a Hilbert $\mathcal{H}$ space such that $\langle R,R \rangle + \langle Q,Q\rangle - \langle R,Q \rangle > 0$. Then the function
$
\rho(R,Q) = \frac{\langle R, Q\rangle}{\langle R,R \rangle + \langle Q,Q\rangle - \langle R,Q \rangle}
$
is a positive definite kernel.
\end{theorem}
The IoU is obtained when $R$ and $Q$ are indicator functions
of the respective regions (because $\langle R, Q \rangle = \int
R(x,y)Q(x,y)\,dx\,dy = |R\cap Q|$). This suggests a simple modification to construct a Soft IoU (SIoU) version of the latter. For a region $R = [x_1,x_2]\times [y_1,y_2]$, the indicator can be written as
$
R(x,y) = H(x - x_1)H(x_2 - x)H(y - y_1)H(y_2-y)
$
where $H(z) = [z \geq 0]$ is the Heaviside step function. SIoU is obtained by replacing the indicator by the smoother function $H_\alpha(z) = \exp(\alpha z) / (1 + \exp(\alpha z))$ instead. Note that SIoU is non-zero even when regions do not intersect.

Theorem~\ref{t:overlap} provides also an interpretation of the geometric embedding $\phi^g$ of eq. \eqref{e:geomembed} as a vector of region coordinates relative to the anchors. In fact, its entries can be written as $\rho(R, R_{p_k,\bx})=\langle \psi_\text{SIoU}(R),
\psi_\text{SIoU}(R_{p_k,\bx}) \rangle$ where $\psi_\text{SIoU}(R)\in \mathcal{H}_\text{SIoU}$ is the linear embedding (feature map) induced by the kernel $\rho$\footnote{The anchor vectors $\psi_\text{SIoU}(R_{p_k,\bx})$ are not necessarily orthonormal (they are if anchors do not overlap), but this can be restored up to a linear transformation of the coordinates.}.

\myparagraph{Modeling multiple aspects} So far, we have assumed that all parts are always visible; however, anchors also
provide a mechanism to deal with the multiple aspects of 3D objects. As depicted in \cref{f:anchors}.c, as the object
rotates out of plane, anchors naturally appear and disappear, therefore activating and de-activating aspect-specific
components in the model. In turn, this allows to model viewpoint-specific parts or appearances. In practice, we extract
the $L$ highest scoring detections $R_{l}$ of the same anchor $p_k$, and keep the one closest to $R$.

In order to allow anchors to turn off in the model, the geometric embedding is modified as follows. Let $s_k(R_{l}|\bx)$ be the detection score of anchor $k$ in correspondence of the region $R_{l}$; then
\begin{equation}\label{e:boxsimilarity}
\boxsim(R,R_{p_k,\bx}) 
= \max_{l \in \{1, \dots, L\}}
\siou(R,R_{l}) \times \max\{0,s_k(R_l|\bx)\}.
\end{equation}
If the anchor is never detected ($s_k(R_l|\bx) \leq 0$ for all $R_l$) then $\boxsim(R,R_{p_k,\bx}) = 0$. Furthermore,
this expression also disambiguates ambiguous anchor detections by picking the one closest to $R$. Note that in eq.~\eqref{e:boxsimilarity} one can still interpret the factors $\siou(R,R_{l})$ as projections $\langle \psi_\text{SIoU}(R),\psi_\text{SIoU}(R_{l}) \rangle$.

\myparagraph{Relation to DPM} DPM is also a MIL method using a joint embedding $\phi^\text{DPM}(\bx|R_1,\dots,R_K)$ that codes simultaneously for the appearance of $K$ parts and their pairwise geometric  relationships. Our Webly-supervised learning problem requires a representation that can bridge object-focused images (where several parts are visible together as components) and part-focused images (where parts are regarded as objects in their own right). This is afforded by our embedding $\phi^{ag}(\bx|R)$ but not by the DPM one. Besides bridging parts as components and parts as objects, our embedding is very robust (important in order to deal with very noisy training labels), automatically codes for multiple object aspects, and bridges unsupervised non-semantic parts (the anchors) with semantic ones.

\subsection{Anchors: weakly-supervised non-semantic parts}\label{s:mid-level}

\newcommand{\bomega}{\boldsymbol{\omega}}

The geometric embedding in the previous section leverages the power of an intermediate representation: a collection of anchors $\{p_k\}_{k=1}^K$, learned automatically using weak supervision. While there are many methods to discover discriminative non-semantic mid-level parts from image collections (\cref{s:related}), here we propose a simple alternative that, empirically, works better in our context.

We learn the anchors using a formulation similar to the MIL objective (eq.~\eqref{e:objective1}):
\begin{equation}\label{eq:anchors}
\min_{\bomega_1,\dots,\bomega_K}
\sum_{i=1}^K
\left[
\frac{\lambda}{2}
\|\bomega_k\|^2
-
\frac{1}{n} \sum_{i=1}^n  y_i \left[ \max_{R\in\mathcal{R}(\bx_i)}\langle \phi^a(\bx_i|R),\bomega_k\rangle \right]_{+}
\right]
+
\gamma \sum_{k\not= q}
 \left\langle
 \frac{\bomega_k}{\|\bomega_k\|},
 \frac{\bomega_q}{\|\bomega_q\|}
 \right\rangle^2,
\end{equation}
where $[z]_+ = \max\{0,z\}$. Intuitively, anchors are learnt as discriminative
mid-level parts using weak supervision. Anchor scores $s_k(R|\bx) = \langle
\phi^a(\bx|R),\bomega_k\rangle$ are parametrized by vectors
$\bomega_1,\dots,\bomega_K$; the first term in \cref{eq:anchors} is akin to the
baseline MIL formulation of \cref{s:mil} and encourages each anchor $p_k$ to score highly in images $\bx_i$ that contain the object ($y_i = +1$) and to be inactive otherwise ($y_i = -1$). The last term is very important and encourages the learned models $\{\bomega_k\}_{k=1}^K$ to be mutually orthogonal, enforcing \emph{diversity}. Note that anchors use the pure appearance-based region descriptor $\phi^a(\bx)$ since the geometric-aware descriptor $\phi^{ag}(\bx)$ can be computed only once anchors are available. Optimization uses stochastic gradient descent with momentum.

This formulation is similar to the MIL approach of \cite{jian2015learning} which, however, does not contain the orthogonality term. When this term is removed, we observed that the solution degenerates to detecting the most prominent object in an image. \cite{doersch2013mid} uses instead a significantly more complex formulation inspired by mode seeking; in practice we opted for our approach due to its simplicity and effectiveness.

\subsection{Incorporating strong annotations in MIL}\label{s:mil-single}

While we are primarily interested in understanding whether semantic object parts can be learned from Web sources alone, in some cases the precise definition of the extent of a part is inherently ambiguous (\eg~what is the extent of a ``human nose''?). Different benchmark datasets may use somewhat different definition of these concepts, making evaluation difficult. In order to remove or at least reduce this dataset-dependent ambiguity, we also explore the idea of using a single strongly annotated example to fix this degree of freedom.

Denote by $(\bx_a,R_a)$ the single strongly-annotated example of the target
part. This is incorporated in the MIL formulation, eq.~\eqref{e:objective1}, by augmenting the score with a factor that compares the appearance of a region to that of $R_a$:
\begin{equation}
\label{eq:single}
\langle \phi(\bx_i|R),\bw\rangle 
\times
\begin{cases}
\frac{1}{C}\exp \beta \langle \phi^a(\bx_i|R), \phi^a(\bx_a|R_a) \rangle,
& y_i = +1, \\
1, & y_i = -1. 	
\end{cases}
\end{equation}
where
$
C = 
\operatornamewithlimits{avg}_{i : y_i = + 1}
\exp 
\beta \langle \phi^a(\bx_i|R), \phi^a(\bx_a|R_a) \rangle
$
is a normalizing constant. In practice, this is used only during
re-localization rounds of the training phase to guide spatial selection; at test
time, bounding boxes are scored solely by the model of eq.~\eqref{e:objective1} without the additional term. Other formulations, that may use a mixture of strongly and Webly supervised examples, are also possible. However, this is besides our focus, which is to see whether parts are learnable from the Web automatically, and the single supervision is only meant to reduce the ambiguity in the task for evaluation.

\section{Experiments}\label{s:experiments}

This section thoroughly evaluates the proposed method. Our main evaluation is a comparison with existing state-of-the-art techniques on the task of Webly-supervised semantic part learning. In \cref{s:exp-detector} we show that our method is substantially more accurate than existing alternatives and, in some cases, close to fully-supervised part learning.

Having established that, we then evaluate the weakly-supervised mid-level part learning (\cref{s:mid-level})
that is an essential part of our approach. It compares favorably in terms of simplicity, scalability, and accuracy against existing
alternatives for discriminability as well as spatial matching of object categories (\cref{s:exp-matching}).

\myparagraph{Datasets}
The Labeled Face Parts in the Wild (LFPW) dataset \cite{belhumeur2013localizing} contains about 1200 face
images annotated with outlines for landmarks. Outlines are converted into bounding box annotations and images with
missing annotations are removed from the test set. These test images are used to locate the following
entities: \textit{face, eye, eyebrow, nose}, and \textit{mouth}.

The PascalParts dataset \cite{chen14detect} augments the PASCAL VOC 2010 dataset with segmentation masks for
object parts. Segmentation masks are converted into bounding boxes for evaluation. Parts of the same type (\eg~left and right wheels) are merged in a single entity (\textit{wheel}). Objects marked as truncated or difficult are not considered for evaluation. The evaluation focuses on the bus and car categories with 18 entity types overall: \textit{car}, \textit{bus}, and their \textit{door}, \textit{front}, \textit{headlight},
\textit{mirror}, \textit{rear}, \textit{side}, \textit{wheel}, and \textit{window} parts. This dataset is more
challenging, as entities have large intra-class appearance and pose
variations. The evaluation is performed on images from the validation set that contain at least one object instance. 
Furthermore, following \cite{wang15joint}, object occurrences are roughly localized before detecting the parts using their
localization procedure.
Finally, objects whose bounding box larger side is smaller than 80 pixels are removed as several parts are nearly invisible below that scale.

The training sets from both datasets are utilized solely for training the fully supervised baselines
(\cref{s:exp-detector}), and they are not used by MIL approaches.

\myparagraph{Experimental details}
Regions are extracted using selective search \cite{uijlings13selective}, and described using $\ell_2$-normalized Decaf \cite{jia2014caffe} fc6 features to compute the appearance embedding $\phi^a(\bx|R)$. The context descriptor $\mu(R)$ is extracted from a region double the size of $R$. The joint appearance-geometric embedding $\phi^{ag}(\bx|R)$ is obtained by first extracting the top $L=5$ non-overlapping detections of each anchor and then applying \cref{e:boxsimilarity,e:agkron}.

A separate mid-level anchor dictionary $\{p_1, ..., p_K\}$ is learnt for each
object class using the Web images for all the semantic parts for the target
object (including images of the object as a whole) as positive images and the
background clutter images of \cite{fei2007learning} as negative ones. Eq. (\ref{eq:anchors}) is optimized using stochastic gradient descend (SGD) with momentum for 40k iterations, alternating between positive and negative images. We train 150 anchor detectors per object class.

MIL semantic part detectors are trained solely on the Web images and the background class of \cite{fei2007learning} is used as
negative bag for all the objects. The first five relocalization rounds are performed using the appearance only
and the following five use the joint appearance-geometry descriptor (the joint embedding performs better with these two
distinct steps). The MIL $\lambda$ hyperparameter is set by performing 
leave-one-category-out cross-validation\footnote{In other words, $\lambda$ is validated on the training sets of two
  object classes; the best parameter setting is then applied to the
  remaining class, for which strong annotations remain unavailable.}.

Web images for parts are acquired by querying the BING image search engine. For car and bus parts, the query 
concatenates the object and the part names (e.g. "car door"). For face parts,
we do not use the object name. We retrieve 500 images of the class corresponding to the object itself and 
100 images of all other semantic part classes.

\subsection{Webly supervised localization of objects and semantic parts}
\label{s:exp-detector}

This section evaluates the detection performance of our approach. 
We gradually incorporate the proposed improvements, \ie~the context descriptor (C) and the geometrical embedding (G) to
the basic MIL baseline (B) as defined in \cref{s:mil} and monitor their impact.

We compare our method to the \sota co-localization algorithm of
Cho~\etal~\cite{cho15unsupervised} and the \sota weakly supervised detection
method from Bilen and Vedaldi \cite{bilen2015wsdnn}. To detect a given 
part with \cite{cho15unsupervised}, we run their code on all images that contain
that part (\eg~for co-localizing eyes we consider \textit{face} and \textit{eye}
images).
As reference, we also report a fully supervised detector, trained using bounding-boxes
from the training set, for all objects and parts (F).
For this, we use the R-CNN method of
\cite{girshick2014rich} on top of the same features used in MIL. 

We mainly report the Average Precision (AP) per
part/object class and its average (mAP) over all parts in each class.
We also report the CorLoc (for correct localization) measure, as it is often used in the
co-localization literature \cite{desealers2010localizing,joulin14efficient}. As most parts in both datasets are
relatively small, following \cite{chen14detect}, the $\iou$ threshold for correct detection is set to 0.4.

\begin{table}[t!]
\centering 
\begin{tabular}{|l|ccc|ccc|} \hline
 measure          & \multicolumn{3}{c|}{mAP}                                      & \multicolumn{3}{c|}{averageCorLoc}                            \\ \hline
 Parent class  &  \{face\}               & \{car\}  & \{bus\}    & \{face\}          & \{car\}                & \{bus\}                          \\ \hline
Cho \etal~\cite{cho15unsupervised}  & 16.6               & 16.9               & 12.4               & 31.4         & 29.9 & 15.5               \\
Bilen \& Vedaldi~\cite{bilen2015wsdnn}  & 2.7                & 12.0               & 4.7                & 7.2      & 15.3 & 6.7                \\
\cline{1-1}
B      & 20.6          & 29.1          & 22.7          & 22.0          & 38.1          & 29.4          \\
B+C    & 22.4          & 27.3          & 21.4          & 29.1          & 37.6          & 28.4          \\
B+G    & 29.0          & 34.1          & \textbf{23.3} & 33.1          & 45.5          & \textbf{31.5} \\
B+C+G  & \textbf{44.9} & \textbf{34.4} & 23.0            & \textbf{52.5} & \textbf{47.8} & 29.6          \\ \hline
F     & 53.7         & 51.2          & 48.2          & 60.5          & 62.9         & 63.8          \\
F+C+G & \textbf{61.4} & \textbf{60.3} & \textbf{54.1} & \textbf{67.8} & \textbf{71.8} & \textbf{66.0} \\ \hline
\end{tabular}
\caption{\textbf{Part detection results averaged} for the face, car, and bus parent classes. mAP and average CorLoc for the MIL baseline (B), our improved versions that use context (C), geometrical
  embedding (G) compared to the fully supervised R-CNN (F).}
\label{tab:detPerf}
\vspace{-0.5cm}
\end{table}

\begin{table}[t!]
\centering
\begin{tabular}{|c|l|m{0.85cm}m{0.85cm}m{0.85cm}m{0.85cm}m{0.85cm}m{0.85cm}m{0.85cm}m{0.85cm}m{0.87cm}|c|} \hline
\multicolumn{2}{|c|}{Class}  & door & rear & wheel & wind. & side & car & front& headl. & mirror   & mean\{car\}      \\ \hline
\multirow{4}{*}{Web}  & B     & 0.4           & 10.8          & 34.9          & 3.6           & 63.1          & 92.6          & \textbf{55.2} & 0.7           & \textbf{0.3}  & 29.1          \\
                      & B+C   & 0.8           & 11.4          & 31.3          & 4.9           & 58.8          & 83.0          & 54.0          & \textbf{1.0}  & 0.2           & 27.3          \\
                      & B+G   & 0.7           & 11.8          & \textbf{47.9} & \textbf{22.7} & 71.3          & \textbf{97.8} & 54.5          & 0.2           & 0.2           & 34.1          \\
                      & B+C+G & \textbf{5.1}  & \textbf{14.7} & 43.6          & 22.6          & \textbf{72.3} & 95.7          & 54.7          & 0.3           & 0.2           & \textbf{34.4} \\ \hline
\multirow{2}{*}{Full} & F     & 17.0          & \textbf{39.0} & 66.3          & 53.3          & 83.2          & 95.1          & 75.9          & 25.3          & 5.5           & 51.2          \\
 & F+C+G & \textbf{31.1} & 30.7 & \textbf{72.3} & \textbf{67.3} & \textbf{90.1} & \textbf{98.7} & \textbf{82.9} & \textbf{48.1} & \textbf{21.3} & \textbf{60.3} \\ \hline
\end{tabular}
\caption{\textbf{Individual part detection results for car}: APs for the MIL baseline (B), our improved versions that use context (C), geometrical embedding (G)
  and the different flavors of the fully supervised R-CNN (F).}
\label{tab:detPerfCar}
\end{table}

\myparagraph{Results}
\Cref{tab:detPerf} reports the average AP and CorLoc over all parts of a given object class for all these methods.
First, we observe that even the MIL baseline (B) outperforms
off-the-shelf methods such as \cite{cho15unsupervised} and 
\cite{bilen2015wsdnn}. For \cite{bilen2015wsdnn}, we have observed that the part detectors degrade
to detecting subparts of semantic parts, suggesting that \cite{bilen2015wsdnn} lacks robustness to drastic scale variations and
to the large amount of noise present in our dataset.
Second, we see that using the geometric embedding (+G)
always improves the baseline results by $1-10$ mAP points. On top of geometry, using context (+C)
helps for face and car parts, but not for buses. Overall the unified embedding brings a large improvement for faces
(+24.3 mAP) and for cars (+5.3 mAP) and more contained for buses (+0.6 mAP). Importantly, these improvements
significantly reduce the gap between using noisy Web supervision and the fully supervised R-CNN (F); overall, Webly
supervision achieves respectively 84\%, 67\%, and 48\% of the performance of (F). 

Last but not least, we extended the fully supervised R-CNN method with the joint appearance-geometry embedding and the context descriptor (F+C+G), which improves part detections by +7.7, +9.1, +5.9 mAP points respectively. This suggests that our representation may be applicable well beyond weakly supervised learning.

Table~\ref{tab:detPerfCar} shows per-part detection results for the car parts. We see that geometry helps for 6 parts out of 9. Out of the three remaining parts, two are cases for which the MIL baseline failed completely. In the less ambiguous fully-supervised scenario, the geometric embedding improves the performance in 8 out of 9 cases.

\begin{table}[t!]
\centering 
\begin{tabular}{|l|rrr|rrr|} \hline
 measure          & \multicolumn{3}{c|}{mAP}  & \multicolumn{3}{c|}{averageCorLoc}    \\ \hline
 Parent class  &  \{face\}               & \{car\}  & \{bus\}    & \{face\}          & \{car\}                & \{bus\} \\ \hline
A       & 29.4 $\pm$ 2.6 & 25.1 $\pm$ 2.7 & 24.5 $\pm$ 2.7 & 38.2 $\pm$ 2.5 & 39.8 $\pm$ 3.2 & 39.6 $\pm$ 3.2 \\
A+B     & 27.3 $\pm$ 3.1 & 33.3 $\pm$ 1.1 & 26.9 $\pm$ 1.3 & 34.6 $\pm$ 3.7 & 46.6 $\pm$ 1.5 & 40.0 $\pm$ 2.3 \\
A+B+C   & 38.2 $\pm$ 3.1 & 32.4 $\pm$ 1.2 & 26.6 $\pm$ 1.6 & 51.7 $\pm$ 3.2 & 49.4 $\pm$ 1.5 & 43.9 $\pm$ 3.0 \\
A+B+G   & 34.5 $\pm$ 4.3 & 35.7 $\pm$ 1.1 & 28.1 $\pm$ 1.2 & 43.5 $\pm$ 4.8 & 48.8 $\pm$ 1.6 & 42.2 $\pm$ 2.2 \\
A+B+C+G & \textbf{43.0 $\pm$ 3.6} & \textbf{36.4 $\pm$ 1.0} & \textbf{30.1 $\pm$ 1.8} & \textbf{54.7 $\pm$ 3.2} & \textbf{51.6 $\pm$ 1.6} & \textbf{45.9 $\pm$ 2.8} \\ \hline
\end{tabular}
\caption{\textbf{Part detection results using a single strong annotation} (A): mAP and average CorLoc for the MIL baseline (B), our improved versions that use context (C), geometrical
  embedding (G). Mean and standard deviation over 25 random annotations.}
\label{tab:detPerfSingleAnno}
\end{table}

\myparagraph{Leveraging a single annotation} As noted in \cref{s:mil-single},
one issue with weakly supervised part learning is the inherent ambiguity in the
 part extent, that may differ from dataset to dataset. Here we
address the ambiguity by adding a single strong annotation to the mix using the
method described in section \ref{s:mil-single}.
We asked an annotator to select 25 representative part annotations per part class from the training sets of each dataset. We retrain every part detector for each of the annotations
and report mean and standard deviation of mAP. As a baseline, we also consider an exemplar detector trained using the single annotated example (A).

Results are reported in \cref{tab:detPerfSingleAnno}. Compared to pure Web supervision (B+C+G) in \cref{tab:detPerf}, the single annotation (A+B+C+G) does not help for faces, for which the proposed method was already working very well, but there is a +2 mAP point improvement for cars and +6.8 mAP for buses, which are more challenging. We also note that the complete method (A+B+C+G) is substantially superior to the exemplar detector (A).

\subsection{Validation of weakly-supervised mid-level anchors}\label{s:exp-anchors}
This section validates the mid-level anchors (\cref{s:mid-level}) against alternatives in terms of discriminative information content and its ability of establishing meaningful matches between images, which is a key requirement in our application.

\myparagraph{Discriminative power of anchors}\label{s:exp-classif}
Since most of the existing methods for learning mid-level patches are evaluated in terms of discriminative 
content in a classification setting, we adopt the same protocol here. In particular, we evaluate the anchors as
mid-level patches on the MIT Scene 67 indoor scene classification task~\cite{quattoni2009recognizing}. The pipeline
first learns 50 mid-level anchors for each of the 67 scene classes. Then, similar to \cite{li2015mid}, images are split into spatial
grids (2x2 and 1x1) and described by concatenating the maximum scores attained by each anchor
detector inside each bin of the grid. All the grid descriptors are then concatenated to form a global image descriptor
which is $\ell_2$ normalized. 67 one-vs-rest SVM classifiers are trained on top of these descriptors. To be comparable
with other methods, we consider both Decaf fc6 and VGG-VD fc7 \cite{simonyan2014very} descriptors.
 
\Cref{tab:classification} contains the results of the classification experiment. 
Our weakly-supervised anchors clearly outperform other mid-level element approaches that are not based on CNN features \cite{Juneja13,doersch2013mid,jian2015learning,bossard2014food}.
Among CNN based approaches, our method outperforms the \sota mid-level feature based method from \cite{li2015mid} on both VGG-VD and Decaf features. 
Remarkably, using our part detectors improves over the baseline which uses the global image CNN descriptor (FC) by 13.8
and 8.7 average accuracy points for Decaf and VGG-VD features respectively. 
Compared to other methods which are not based on detecting mid-level elements,
our pipeline outperforms \sota FV-CNN for Decaf features and is inferior for VGG-VD.

\begin{table}[t!]
\resizebox{\textwidth}{!}{
\begin{tabular}{|c||c|c|c|c||c|c|} \hline
method & BoP$^\dagger$ \cite{Juneja13} & DMS$^\dagger$ \cite{doersch2013mid} & Jian \etal$^\dagger$ \cite{jian2015learning} &  RFDC$^\dagger$ \cite{bossard2014food} &
\lnbreakcell{FC\\ \textit{Decaf}} \cite{jia2014caffe} &  \lnbreakcell{FC\\\textit{VGG-VD}} \cite{simonyan2014very}\\ \hline
accuracy (\%) & 46.1         & 64.0         & 58.1      &  54.4    & 57.7        & 68.9  \\ \hline\hline
method &  \lnbreakcell{BoE$^\dagger$\\ \textit{Decaf}} \cite{li2015mid} & \lnbreakcell{ours$^\dagger$\\\textit{Decaf}} &
\multicolumn{1}{c||}{\lnbreakcell{FV-CNN\\\textit{Decaf}} \cite{cimpoi2015deep}} & \multicolumn{1}{c|}{\lnbreakcell{BoE$^\dagger$\\\textit{VGG-VD}} \cite{li2015mid}}   & \lnbreakcell{ours$^\dagger$\\\textit{VGG-VD}} & \lnbreakcell{FV-CNN\\\textit{VGG-VD}} \cite{cimpoi2015deep} \\ 
 \hline
accuracy (\%) &  69.7        & \textbf{71.5}          & \multicolumn{1}{c||}{69.7}       & \multicolumn{1}{c|}{77.6}        & 77.8        & \textbf{81.6}   \\ \hline   
\end{tabular}
}
\caption{Classification results on MIT Scenes \cite{quattoni2009recognizing}. Methods using mid-level elements are
  marked with $^\dagger$. For CNN-based approaches, features rely on Decaf or VGG-VD.}
\label{tab:classification}
\end{table}

\myparagraph{Ability of anchors to establish semantic matches}
\label{s:exp-matching}
The previous experiment assessed favorably the mid-level parts in terms of discriminative content; however, in the
embedding $\phi^g$, these are used as \emph{geometric anchors}. Hence, here we validate the ability of the mid-level anchors to induce good semantic matches between pairs of images (before learning the semantic part models). 

To perform semantic matching between a source image $x_S$ and a target image
$x_T$, we consider each part annotation $R_S$ in the source  and
predict its best match $\hat R_T$ in the target. The quality of the match is
evaluated by measuring the IoU between the predicted $\hat R_T$ and ground-truth $R_T$ part. When a part appears more than once (\eg~eyes often appear twice),
we choose the most overlaping pair.
Performance is reported by averaging the match IoU for all part occurrences and pairs of images in the test set, reporting the results for each object category.

Given a source part $R_S$, the joint
appearance-geometry embedding (\textit{anchor-ag}) is extracted for the source part $\phi^{ag}(\bx_S|R_S)$ and the target region $\hat R_T$ that maximizes the inner product
$
\langle \phi^{ag}(\bx_S|R_S),
\phi^{ag}(\bx_T|\hat R_{T}) 
\rangle
$
is returned as the predicted match. We also compare \emph{anchor-g} that uses only the geometric embedding
$\phi^g(\bx|R)$ and the baseline \emph{a} that uses only the appearance embedding $\phi^a(\bx|R)$.

We also compare two strong off-the-shelf baselines: DSP
\cite{kim13deformable}, \sota pairwise semantic matching method, and the method of \cite{zhou15flowweb}, \sota for joint alignment.  
To perform box matching with \cite{kim13deformable} and \cite{zhou15flowweb} we fit an affine transformation to the disparity map contained inside
a given source bounding box and apply this transform to move this box to the target image.
Due to scalability issues, we were unable to apply \cite{zhou15flowweb} to the
full dataset\footnote{More precisely, we were not able to apply
  \cite{zhou15flowweb} on a dataset with more than 60 $128\times 68$ pixel images on a
  server with 120 GB of RAM.}, so we perform this comparison on a random subset of 50 images.

\Cref{tab:matching} presents the results of our benchmark. 
On the small subset of 50 images the costly approach of
\cite{zhou15flowweb} performs better than our embedding only on the LFPW faces, where the viewpoint variation is limited. On the car and bus categories our method outperforms \cite{zhou15flowweb} by 10\% and 16\% average IoU respectively. Our method is also consistently better than DSP \cite{kim13deformable} on both the small and full test set.
We also note that the matching using geometric embeddings alone (\textit{anchor-g}) achieves similar performance than the appearance-geometry matching (\textit{anchor-ag}) which validates our intuition that the local geometry of an object is well-captured by the anchors.

\begin{table}[t!]
  \centering
\begin{tabular}{| c  c | c | c | c | c | c|} \hline
                        &      & \multicolumn{5}{c|}{Matching method} \\ 
\cline{3-7}
Set                 & Parent class & anchor-ag      & anchor-g       & \hspace{0.5cm} a \hspace{0.5cm} & Flowweb \cite{zhou15flowweb}        & DSP \cite{kim13deformable}   \\ \hline
\multirow{3}{*}{50 images}  & \{car\}  & \textbf{0.36} & \textbf{0.36} & 0.31 & 0.34          & 0.23 \\ 
                        & \{bus\}  & \textbf{0.37} & 0.36          & 0.31 & 0.31          & 0.22 \\
                        & \{face\} & 0.41          & 0.39          & 0.33 & \textbf{0.43} & 0.19 \\ \hline
\multirow{3}{*}{Full}   & \{car\}  & \textbf{0.36} & \textbf{0.36} & 0.30 & -              & 0.22 \\ 
                        & \{bus\}  & \textbf{0.35} & \textbf{0.35} & 0.29 & -              & 0.21 \\
                        & \{face\} & \textbf{0.41} & 0.39          & 0.34 & -              & 0.21 \\ \hline    
\end{tabular}
\caption{Semantic matching. For every parent class, we report average overlap (IoU) over all semantic parts. The face class results are obtained on the LFPW dataset
while bus and car results come from the PascalParts dataset.}
\label{tab:matching}
\end{table}

\section{An atlas for visual semantic}
\label{s:exp-qualitative} 

\begin{figure}[t!]
  \begin{center}
    \includegraphics[width=\linewidth]{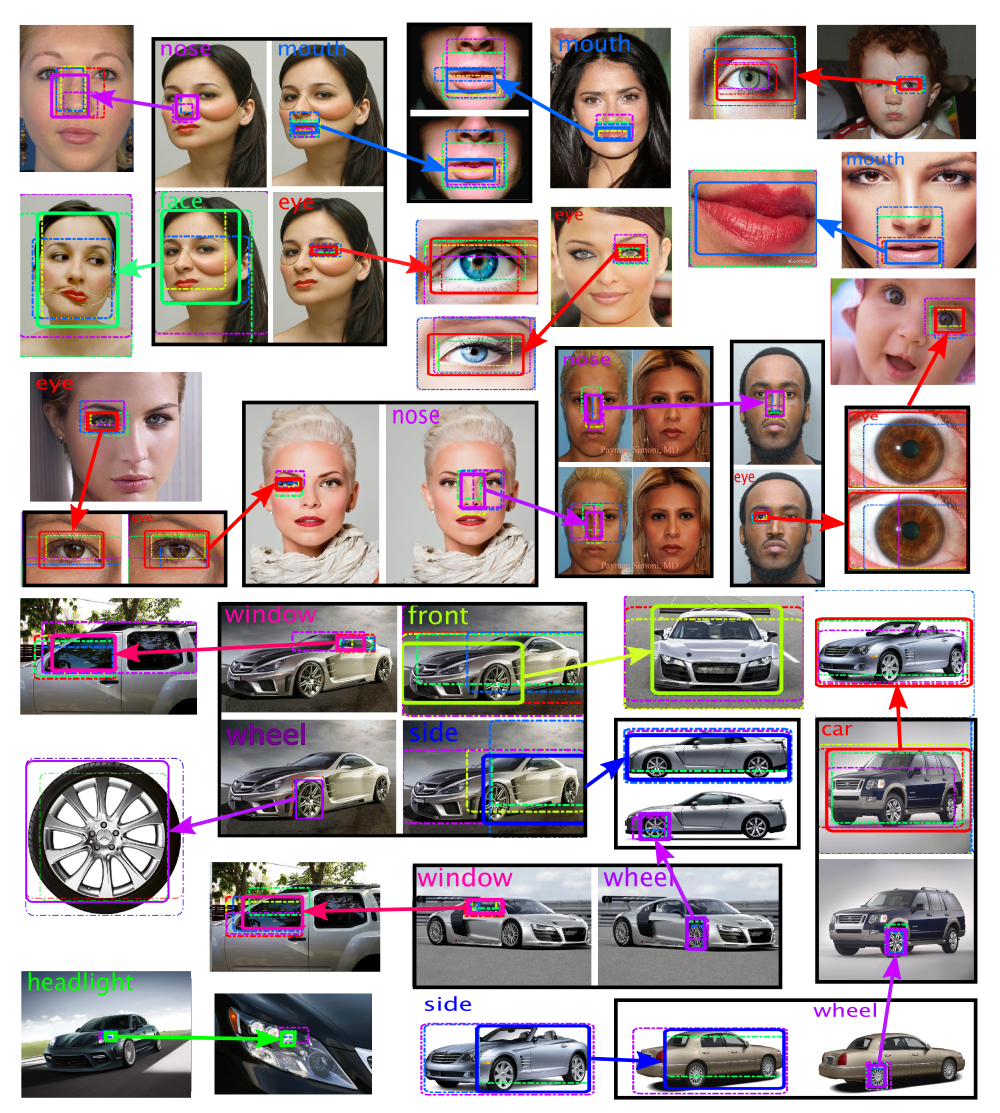}
  \end{center}
\vspace{-0.8cm}
\caption{Navigating the \textit{visual semantic atlas}. Each pair of
solid bounding boxes connected by an arrow denotes a preselected part bounding box
(near the starting point of an arrow) as \textit{detected by our algorithm} and the 
most similar semantic match (the endpoint of the arrow). The best matching bounding box
is the detection with highest appearance-geometry descriptor similarity
among \textit{all the detections in our database} of web images. The dashed boxes denote anchors 
that contributed the most to the similarity.
Please note that the matching gracefully occurs across scales.}
\label{fig:atlas}
\end{figure}

As a byproduct of Webly-supervised learning, our method annotates the Web images with semantic parts. By endowing an
image dataset with such concepts, we show here that it is possible to browse these annotated images. All of this composes
our visual semantic atlas (see a subset of the atlas in Figure \ref{fig:atlas}) that allows to navigate from one image
to another, even between an image of a full object and a zoomed-in image of one of its parts.

\section{Conclusions}
We have proposed a novel method for learning about objects, their semantic parts, and their geometric relationships,
from noisy Web supervision. This is achieved by first learning a weakly supervised dictionary of mid-level visual elements
which define a robust object-centric coordinate frame.
Such property theoretically motivates our approach.
The geometric projections are then used in a novel appearance-geometry embedding that improves learning of semantic object parts from noisy Web data.
We showed improved performance over co-localization \cite{cho15unsupervised}, 
deep weakly supervised approach \cite{bilen2015wsdnn} and a MIL baseline on all benchmarked datasets.
Extensive evaluation of our proposed mid-level elements shows comparable results to \sota in terms of their
discriminative power and superior results in terms of the ability to establish semantic matches between images. 
Finally, our method also provides a visually intuitive way to navigate Web images and predicted annotations.

\paragraph*{Acknowledgments.}

We are grateful for support by XRCE and ERC StG 638009-IDIU.

\bibliographystyle{splncs}
\bibliography{part_transfer}

\appendix
\section{Appendix}\label{s:proofs}

\begin{proof}[Proof of Theorem 1]
The function $\langle R, Q \rangle$ is the linear kernel, which is PD. This kernel is multiplied by the factor $-1/\bar k$ where $\bar k(R,Q)=\langle R,Q \rangle -  \langle R,R \rangle - \langle Q,Q\rangle$; if this factor is also a PD kernel, then the result holds as the product of PD kernels is PD. According to Lemma~3.2 of~\cite{hein05hilbertian}, $-1/\bar k$ is PD if, and only if, $\bar k$ is strictly negative (point-wise) and conditionally definite positive (CDP). The first condition is part of the assumptions. To show the second condition that $\bar k$ is CDP pick $n$ vectors $R_1,\dots,R_n$ and real numbers $c_1,\dots,c_n$ summing to zero $c_1+\dots+c_n=0$; then
\[
\sum_{ij} c_i \bar k(R_i,Q_i) c_j = \sum_{ij} c_i \langle R_i,Q_j \rangle c_j \geq 0
\]
where we used the fact that the terms $\langle R_i,R_i \rangle$  cancel out and the fact that $\langle R_i, Q_j \rangle$ is PD.
\end{proof}

\end{document}